\documentclass[10pt,journal,compsoc]{IEEEtran}

\ifCLASSOPTIONcompsoc
  \usepackage[nocompress]{cite}
\else
  \usepackage{cite}
\fi

\usepackage{times}
\usepackage{graphicx}
\usepackage{amsmath}
\usepackage{amssymb}
\usepackage{soul}
\usepackage{booktabs}

\usepackage{subfig}
\usepackage{url}
\usepackage[utf8]{inputenc}
\ifCLASSINFOpdf
\else
\fi

\begin{document}

\title{Learnable pooling with Context Gating for \\video classification}

\author{Antoine Miech,
        Ivan Laptev
        and Josef Sivic\\
        \small \url{https://github.com/antoine77340/LOUPE}
\IEEEcompsocitemizethanks{\IEEEcompsocthanksitem A. Miech, I. Laptev and J. Sivic
are with Inria, WILLOW, Departement d'Informatique de l'École Normale Supérieure, PSL Research University, ENS/INRIA/CNRS UMR 8548, Paris, France\protect\\
E-mail: \{antoine.miech, ivan.laptev, josef.sivic\}@inria.fr
\IEEEcompsocthanksitem J. Sivic is also with Czech Institute of Informatics, Robotics and Cybernetics, Czech Technical University in Prague.}
}

\markboth{}%
{Shell \MakeLowercase{\textit{et al.}}: Learnable pooling with Context Gating for video classification}
%



\IEEEtitleabstractindextext{%
\begin{abstract}

Current methods for video analysis often extract frame-level features using pre-trained convolutional neural networks (CNNs).
%
Such features are then aggregated over time e.g., by simple temporal averaging or more sophisticated recurrent neural networks such as long short-term memory (LSTM) or gated recurrent units (GRU).
In this work we revise existing video representations and study alternative methods for temporal aggregation.
We first explore clustering-based aggregation layers and propose a two-stream architecture aggregating audio and visual features.
We then introduce a learnable non-linear unit, named Context Gating, aiming to model interdependencies among network activations.
Our experimental results show the advantage of both improvements for the task of video classification. 
In particular, we evaluate our method on the large-scale multi-modal Youtube-8M v2 dataset and outperform all other methods in the Youtube 8M Large-Scale Video Understanding challenge.

\end{abstract}

}

\maketitle

\IEEEdisplaynontitleabstractindextext

%
\IEEEpeerreviewmaketitle

\begin{figure}[t]
  \begin{center}
     \includegraphics[width=0.5\textwidth]{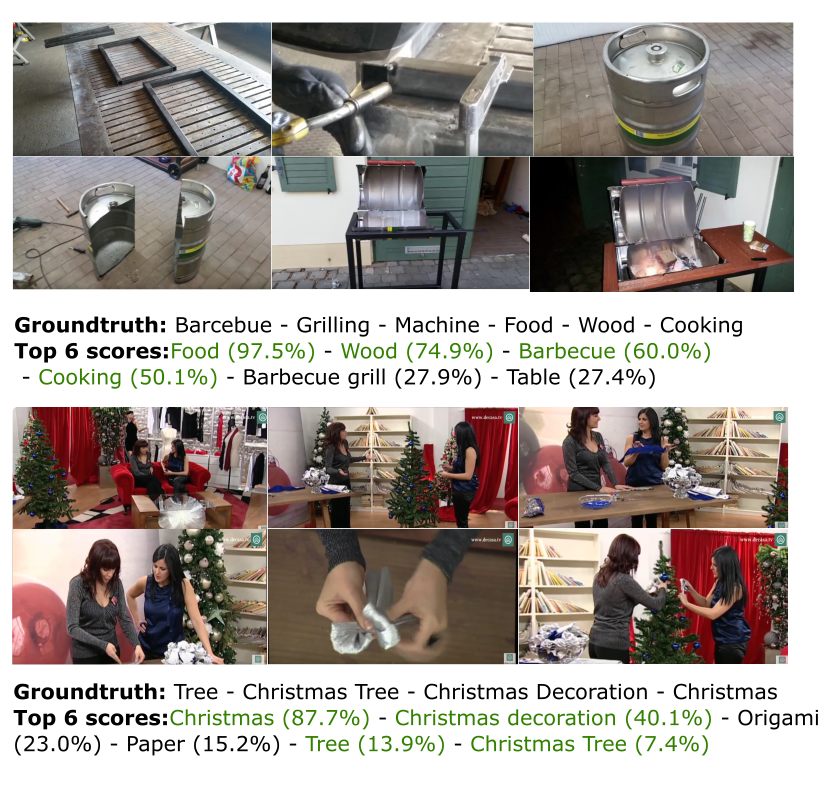}
\end{center}
\vspace*{-3mm}
\caption{Two example videos from the Youtube-8M V2 dataset together with the ground truth and top predicted labels. Predictions colored as green are labels from the groundtruth annotation.}
\label{fig:teaser}
\end{figure}

\IEEEraisesectionheading{\section{Introduction}\label{sec:introduction}}
Understanding and recognizing video content is a major challenge for numerous applications 
including surveillance, personal assistance, smart homes, autonomous driving, stock footage search and sports video analysis.
In this work, we address the problem of multi-label video classification for user-generated videos on the Internet.
The analysis of such data involves several challenges.
Internet videos have a great variability in terms of content and quality (see Figure~\ref{fig:teaser}).
Moreover, user-generated labels are typically incomplete, ambiguous and may contain errors.


Current approaches for video analysis typically represent videos by features extracted from consecutive frames, followed by feature aggregation over time. 
Example methods for feature extraction include deep convolutional neural networks (CNNs) pre-trained on static images~\cite{he16resnet,krizhevsky12imagenet,simonyan2014vgg,szegedy16inception}.
Representations of motion and appearance can be obtained from CNNs pre-trained for video frames and short video clips~\cite{tran15c3d,feichtenhofer16convolutionaltwostream}, as well as hand-crafted video features~\cite{laptev08learning,schuldt04recognizing,wang13action}. 
Other more advanced models employ hierarchical spatio-temporal convolutional architectures~\cite{baccouche11sequentialdeep,carreira2017quovadis,feichtenhofer17spatiotemporalmultiplier,ji133dcnn,tran15c3d,varol17longterm} to both extract and temporally aggregate video features at the same time.

Common methods for temporal feature aggregation include simple averaging or maximum pooling as well as more sophisticated pooling techniques such as VLAD~\cite{jegou10vlad} or more recently recurrent models (LSTM~\cite{hochreiter97lstm} and GRU~\cite{cho11GRU}). These techniques, however, may be suboptimal. Indeed, simple techniques such as average or maximum pooling may become inaccurate for long sequences. Recurrent models are frequently used for temporal aggregation of variable-length sequences~\cite{donahue14long,youtube8m}
and often outperform simpler aggregation methods, however, their training remains cumbersome.
As we show in Section~\ref{exp}, training recurrent models requires relatively large amount of data. 
Moreover, recurrent models can be sub-optimal for processing of long video sequences during GPU training.
It is also not clear if current models for sequential aggregation are well-adapted for video representation.
Indeed, our experiments with training recurrent models using temporally-ordered and randomly-ordered video frames show similar results.

Another research direction is to exploit traditional orderless aggregation techniques based on clustering approaches such as Bag-of-visual-words~\cite{csurka04visual,sivic03videogoogle}, Vector of Locally aggregated Descriptors (VLAD)~\cite{jegou10vlad} or Fisher Vectors~\cite{perronnin12fisher}. It has been recently shown that integrating VLAD as a differentiable module in a neural network can significantly improve the aggregated representation for the task of place retrieval~\cite{arandjelovic16netvlad}. This has motivated us to integrate and enhance such clustering-based aggregation techniques for the task of video representation and classification.

\hfill \break
{\bf Contributions.} 
In this work we make the following contributions:
(i)~we introduce a new state-of-the-art architecture aggregating video and audio features for video classification,
(ii)~we introduce the Context Gating layer, an efficient non-linear unit for modeling interdependencies among network activations, and
(iii)~we experimentally demonstrate benifits of clustering-based aggregation techniques over LSTM and GRU approaches for the task of video classification. 

\hfill \break
{\bf Results.} 
We evaluate our method on the large-scale multi-modal Youtube-8M V2 dataset containing about 8M videos and 4716 unique tags.
We use pre-extracted visual and audio features provided with the dataset~\cite{youtube8m} and demonstrate improvements obtained with the Context Gating as well as by the combination of learnable poolings.
Our method obtains top performance, out of more than 650 teams, in the Youtube-8M Large-Scale Video Understanding challenge\footnote{\url{https://www.kaggle.com/c/youtube8m}}.
Compared to the common recurrent models, our models are faster to train and require less training data.
Figure~\ref{fig:teaser} illustrates some qualitative results of our method.

\section{Related work}
This work is related to previous methods for video feature extraction, aggregation and gating reviewed below.


\subsection{Feature extraction}
Successful hand-crafted representations~\cite{laptev08learning,schuldt04recognizing,wang13action} are 
based on local histograms of image and motion gradient orientations extracted along dense trajectories~\cite{souza16symathy,wang13action}.
More recent methods extract deep convolutional neural network activations computed from individual frames or 
blocks of frames using spatial~\cite{feichtenhofer16convolutionaltwostream,karpathy2014,girdhar17actionvlad,wang2015action} or 
spatio-temporal~\cite{baccouche11sequentialdeep,carreira2017quovadis,feichtenhofer17spatiotemporalmultiplier,ji133dcnn,tran15c3d,varol17longterm} convolutions.
Convolutional neural networks can be also applied separately on the appearance channel and the 
pre-computed motion field channel resulting in the, so called, two-stream 
representations~\cite{carreira2017quovadis,feichtenhofer16convolutionaltwostream,girdhar17actionvlad,simonyan2014,varol17longterm}.  
As our work is motivated by the Youtube-8M large-scale video understanding challenge~\cite{youtube8m}, we will assume for the rest of the paper that features are provided (more details are provided in Section~\ref{exp}). This work mainly focuses on the temporal aggregation of given features.

\subsection{Feature aggregation}
Video features are typically extracted from individual frames or short video clips.
The remaining question is: how to aggregate video features over the entire and potentially long video? 
One way to achieve this is to employ recurrent neural networks, such as long short-term memory (LSTM)~\cite{hochreiter97lstm} or  gated recurrent unit (GRU)~\cite{cho11GRU}), on top of the extracted frame-level features to capture the temporal 
structure of video into a single representation~\cite{fernando15modeling,donahue14long,ibrahim16hierarchical,lev16rnnfisher,yue15beyondshort}.
Hierarchical spatio-temporal convolution architectures~\cite{baccouche11sequentialdeep,carreira2017quovadis,feichtenhofer17spatiotemporalmultiplier,ji133dcnn,tran15c3d,varol17longterm} can also be viewed as a way to both extract and aggregate temporal features at the same time.
Other methods capture only the {\em distribution} of features in the video, not explicitly modeling their temporal ordering.
The simplest form of this approach is the average or maximum pooling of video features~\cite{wang16temporal} over time. 
Other commonly used methods include bag-of-visual-words~\cite{csurka04visual,sivic03videogoogle}, Vector of Locally aggregated Descriptors (VLAD)~\cite{jegou10vlad} or Fisher Vector~\cite{perronnin12fisher} encoding. 
Application of these techniques to video include~\cite{laptev08learning,peng14boosting,schuldt04recognizing,wang13action,xu15discriminative}. 
Typically, these methods~\cite{lev16rnnfisher,perronnin15fisher} rely on an unsupervised learning of the codebook. However, the codebook can also be learned in a discriminative manner~\cite{peng14boosting,peng14stackedfv,sydorov14deepfisher}
or the entire encoding module can be included within the convolutional neural network architecture and trained in the end-to-end manner~\cite{arandjelovic16netvlad}.  
This type of end-to-end trainable orderless aggregation has been recently applied to video frames in~\cite{girdhar17actionvlad}. Here we extend this work by aggregating visual and audio inputs, and also investigate multiple orderless aggregations.

\subsection{Gating}

Gating mechanisms allow multiplicative interaction between a given input feature $X$ and a gate vector with values in between 0 and 1. They are commonly used in recurrent neural network models such as LSTM~\cite{hochreiter97lstm} and GRU~\cite{cho11GRU} but have so far not been exploited in conjunction with other non-temporal aggregation strategies such as Fisher Vectors (FV), Vector of Locally Aggregated Descriptors (VLAD) or bag-of-visual-words (BoW). Our work aims to fill this gap and designs a video classification architecture combining non-temporal aggregation with gating mechanisms.  One of the motivations for this choice is the recent Gated Linear Unit (GLU)~\cite{dauphin16GLU}, which has demonstrated  significant improvements in natural language processing tasks. 

Our gating mechanism initially reported in~\cite{miech17learnable} is also related to the parallel work on Squeeze-and-Excitation architectures~\cite{hu17squeezeandexcitation}, that has suggested gated blocks for image classification tasks and have demonstrated excellent performance on the ILSVRC 2017 image classification challenge. 

\begin{figure}[t]
  \begin{center}
     \includegraphics[width=0.5\textwidth]{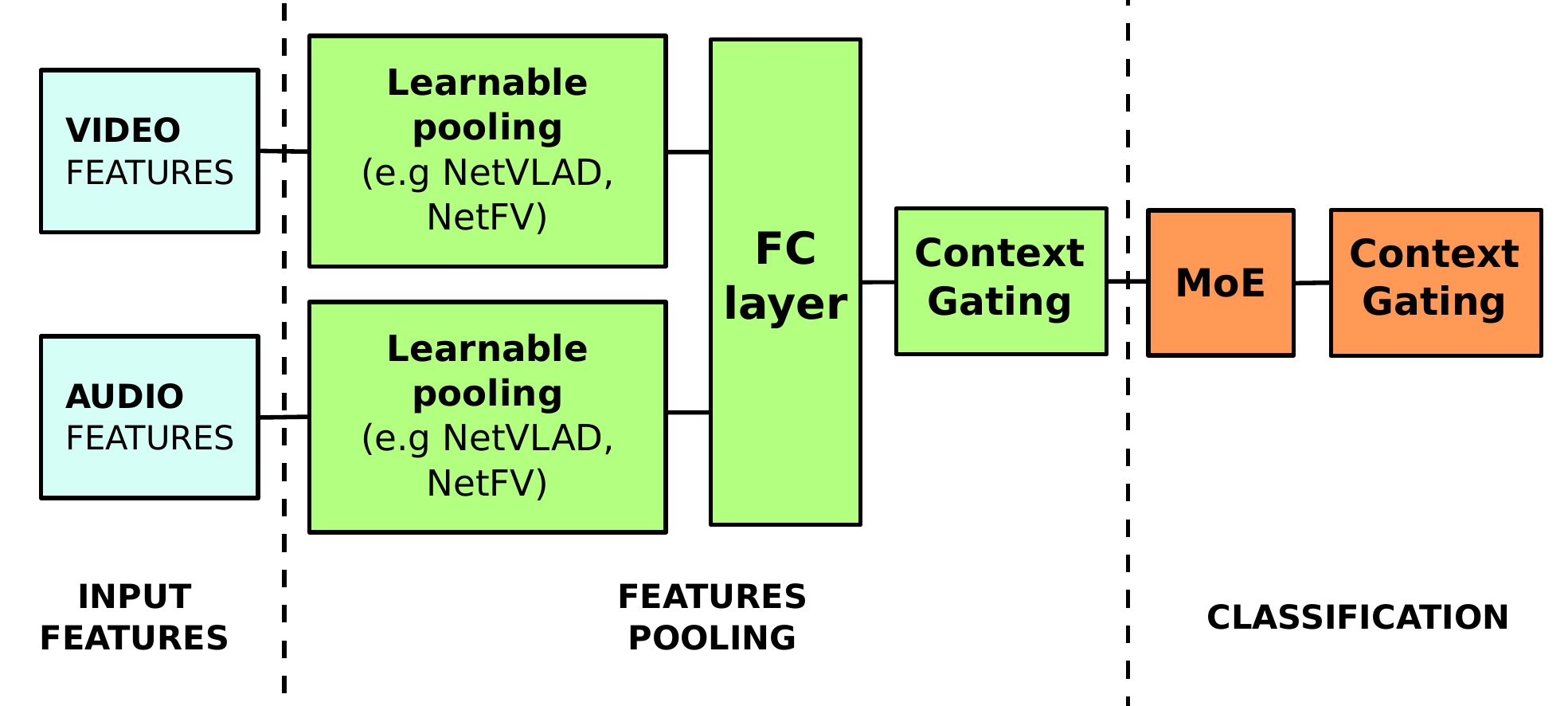}
\end{center}
\caption{Overview of our network architecture for video classification (the ``Late Concat" variant). 
FC denotes a Fully-Connected layer. MoE denotes the Mixture-of-Experts classifier~\cite{youtube8m}.}
\label{fig:model}
\end{figure}

\section{Network architecture}
\label{sec:arch}

Our architecture for video classification is illustrated in Figure~\ref{fig:model} and contains three main modules. First, the input features are extracted from video and audio signals. Next, the pooling module aggregates the extracted features into a single compact (e.g. 1024-dimensional) representation for the entire video.
This pooling module has a two-stream architecture treating visual and audio features separately. The aggregated representation is then enhanced by the Context Gating layer (section \ref{sec:cg}).
Finally, the classification module takes the resulting representation as input and outputs scores for a pre-defined set of labels. The classification module adopts the Mixture-of-Experts~\cite{jordan94mixture} classifier as described in~\cite{youtube8m}, followed by another Context Gating layer.

\subsection{Context Gating} \label{sec:cg}

The Context Gating (CG) module transforms the input feature representation $X$ into a new representation $Y$ as

\begin{align}
\label{eq:cg}
 Y = \sigma(WX+b) \circ X,
\end{align} 
where $X \in \mathbb{R}^{n}$ is the input feature vector, $\sigma$ is the element-wise sigmoid activation and $\circ$ is the element-wise multiplication. $W \in \mathbb{R}^{n \times n}$ and $b \in \mathbb{R}^{n}$ are trainable parameters. The vector of weights $\sigma(WX+b) \in [0, 1]$ represents a set of learned gates applied to the individual dimensions of the input feature $X$. 

The motivation behind this transformation is two-fold. 
First, we wish to introduce non-linear interactions among activations of the input representation. 
Second, we wish to recalibrate the strengths of different activations of the input representation through a self-gating mechanism.
The form of the Context Gating layer is inspired by the Gated Linear Unit (GLU) introduced recently for language modeling~\cite{dauphin16GLU} that considers a more complex class of transformations given by $\sigma(W_{1}X+b_{1}) \circ (W_{2}X+b_{2})$, with two sets of learnable parameters $W_1$, $b_1$ and $W_2$, $b_2$.
Compared to the the Gated Linear Unit~\cite{dauphin16GLU}, our Context Gating in~\eqref{eq:cg} (i)~reduces the number of learned parameters as only one set of weights is learnt, and (ii) re-weights directly the input vector $X$ (instead of its linear transformation) and hence is suitable for situations where $X$ has a specific meaning, such the score of a class label, that is preserved by the layer.    
As shown in Figure~\ref{fig:model}, we use Context Gating in the feature pooling and classification modules. First, we use CG to transform the feature vector before passing it to the classification module. Second, we use CG after the classification layer to capture the prior structure of the output label space. Details are provided below.

\subsection{Relation to residual connections} 

Residual connections have been introduced in~\cite{he16resnet}. They demonstrate faster and better training of deep convolutional neural networks as well as better performance for a variety of tasks.
Residual connections can be formulated as
\begin{align}
\label{eq:residual}
 Y = f(WX+b) + X,
\end{align}
where $X$ are the input features, $(W, b)$ the learnable parameters of the linear mapping (or it can be a convolution), $f$ is a non-linearity (typically Rectifier Linear Unit as expressed in~\cite{he16resnet}). One advantage of residual connections is the possibility of gradient propagation directly into $X$ during training, avoiding the vanishing gradient problem. To show this, the gradient of the residual connection can be written as:

\begin{align}
\label{eq:residual-gradient}
 \nabla Y = \nabla(f(WX+b)) + \nabla X.
\end{align} 
One can notice that the gradient $\nabla Y$ is the sum of the gradient of the previous layer $\nabla X$ and the gradient $\nabla(f(WX+b))$. The vanishing gradient problem is overcome thanks to the term $\nabla X$, which allows the gradient to backpropagate directly from $Y$ to $X$
without decreasing in the norm.
A similar effect is observed with Context Gating which has the following 
gradient equation:
\begin{align}
\label{eq:cg-gradient}
 \nabla Y = \nabla(\sigma(WX+b)) \circ X + \sigma(WX+b) \circ \nabla X.
\end{align} 
%
In this case, the term $\nabla X$ is weighted by activations
$\sigma(WX+b)$. Hence, for dimensions where $\sigma(WX+b)$ are close to 1,
gradients are directly propagated from $Y$ to $X$. In contrast, for
values close to 0 the gradient propagation is
vanished. This property is valuable as it allows to stack several non-linear layers and avoid vanishing gradient problems.

\subsection{Motivation for Context Gating}

\begin{figure}[t]
  \begin{center}
     \includegraphics[width=0.4\textwidth]{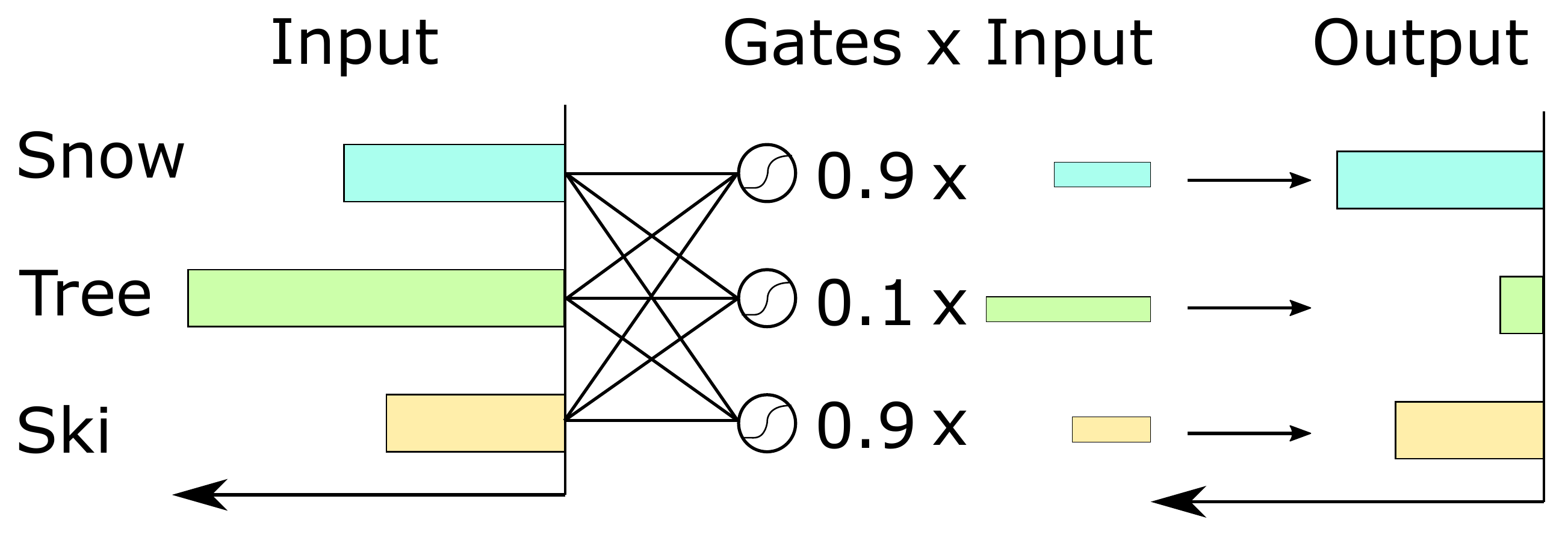}\vspace{-.2cm}\\
\end{center}
\caption{Illustration of Context Gating that down-weights visual activations of Tree for a skiing scene.}
\label{fig:CG-explanation}
\end{figure}

Our goal is to predict human-generated tags for a video. Such tags typically represent only a subset of objects and events which are most relevant to the context of the video. To mimic this behavior and to suppress irrelevant labels, we introduce the Context Gating module both to re-weight the features and the output labels of our architecture.\smallskip\\
%
{\bf \noindent Capturing dependencies among features.} 
Context Gating can help creating dependencies between visual activations. Take an example of a skiing video showing a skiing person, snow and trees. While network activations for the \textit{Tree} features might be high, trees might be less important in the \textit{context} of skiing where people are more likely to comment about the snow and skiing rather than the forest. Context Gating can learn to down-weight visual activations for \textit{Tree} when it co-occurs with  visual activations for \textit{Ski} and \textit{Snow} as illustrated in Figure~\ref{fig:CG-explanation}.
\smallskip\\
%
{\bf \noindent Capturing prior structure of the output space.} 
Context Gating can also create dependencies among output class scores when applied to the classification layer of the network.
This makes it possible to learn a prior structure on the output probability space, which can be useful in modeling \textit{biases in label annotations}.

\section{Learnable pooling methods} \label{lp}
Within our video classification architecture described above, we investigate several types of learnable pooling models, which we describe next.  
Previous successful approaches~\cite{donahue14long,youtube8m} employed recurrent neural networks such as LSTM or GRU for the encoding of the sequential features. We chose to focus on non-recurrent aggregation techniques. This is motivated by several factors: first, recurrent models are computationally demanding for long temporal sequences as it is not possible to parallelize the sequential computation.
Moreover, it is not clear if treating the aggregation problem as a sequence modeling problem is necessary. As we show in our experiments, there is almost no change in performance if we shuffle the frames in a random order as almost all of the relevant signal relies on the static visual cues. All we actually need to do is to find a way to efficiently \textit{remember} all of the relevant visual cues. We will first review the NetVLAD~\cite{arandjelovic16netvlad} aggregation module and then explain how we can exploit the same idea to imitate Fisher Vector and Bag-of-visual-Words aggregation scheme.

\subsection{NetVLAD aggregation}

The NetVLAD~\cite{arandjelovic16netvlad} architecture has been proposed for place recognition to reproduce the VLAD encoding~\cite{jegou10vlad}, but in a differentiable manner, where the clusters are tuned via backpropagation instead of using k-means clustering. It was then extended to action recognition in video~\cite{girdhar17actionvlad}. 
The main idea behind NetVLAD is to write the descriptor $x_i$ hard assignment to the cluster $k$ as a soft assignment:
\begin{align}
\label{eq:softa}
a_{k}(x_i) = \frac{e^{w_k^\top x_i + b_k}}{\sum_{j=1}^{K}e^{w_j^\top x_i + b_j}}
\end{align}
where $(w_j)_j$ and $(b_j)_j$ are learnable parameters. 
In other words, the soft assignment $a_{k}(x_i)$ of descriptor $x_{i}$ to cluster $k$ measures on a scale from $0$ to $1$ how close the descriptor $x_{i}$ is to cluster $k$. In the hard assignment way, $a_{k}(x_i)$ would be equal to $1$ if $x_{i}$ closest cluster is cluster $k$ and $0$ otherwise.
For the rest of the paper, $a_{k}(x_i)$ will define
soft assignment of descriptor $x_i$ to cluster $k$. If we write $c_j, j \in [1, K]$ the j-th learnable cluster, the NetVLAD
descriptor can be written as
\begin{align}
VLAD(j,k) = \sum_{i=1}^{N} a_{k}(x_i)(x_i(j) - c_k(j)),
\end{align}
which computes the weighted sum of residuals $x_i - c_k$ of descriptors $x_i$  from learnable anchor point $c_{k}$ in cluster $k$.
\subsection{Beyond NetVLAD aggregation}
By exploiting the same cluster soft-assignment idea, we can also imitate similar operations
as the traditional Bag-of-visual-words~\cite{csurka04visual,sivic03videogoogle} and Fisher Vectors~\cite{perronnin12fisher} in a differentiable manner.

For bag-of-visual-words (BOW) encoding, we use soft-assignment of descriptors to visual word clusters~\cite{arandjelovic16netvlad,philbin08lost} to obtain a differentiable representation. 
The differentiable BOW representation can be written as:

\begin{align}
BOW(k) = \sum_{i=1}^{N} a_{k}(x_i).
\end{align}
Notice that the exact bag-of-visual-words formulation is reproduced if we replace the soft assignment values by its hard assignment equivalent.
This formulation is closely related to the Neural BoF formulation~\cite{passalis17learning}, but differs in the way of computing the soft assignment. In detail,~\cite{passalis17learning} performs a softmax operation over the computed L2 distances between the descriptors and the cluster centers, whereas we use soft-assignment given by eq.~(\ref{eq:softa}) where parameters $w$ are learnable without explicit relation to computing L2 distance to cluster centers.  It also relates to~\cite{richard15bow} that uses a recurrent neural network to perform the aggregation.
The advantage of BOW aggregation over NetVLAD is that it aggregates a list of feature descriptors into a much
more compact representation, given a fixed number of clusters. The drawback is that significantly more clusters are needed to obtain a rich representation of the aggregated descriptors.

Inspired by Fisher Vector~\cite{perronnin12fisher} encoding, we also experimented with modifying the NetVLAD architecture to allow learning of second order feature statistics within the clusters.
We will denote this representation as NetFV (for Net Fisher Vectors) as it aims at imitating the standard Fisher Vector encoding~\cite{perronnin12fisher}. Reusing the previously established soft assignment notation, we can write the NetFV representation as
\begin{align}
FV1(j,k) = \sum_{i=1}^{N} a_{k}(x_i)\bigg(\frac{x_i(j) - c_k(j)}{\sigma_k(j)}\bigg),\\
FV2(j,k) = \sum_{i=1}^{N} a_{k}(x_i)\Bigg(\bigg(\frac{x_i(j) - c_k(j)}{\sigma_k(j)}\bigg)^{2} - 1\Bigg),
\end{align}
where $FV1$ is capturing the first-order statistics, $FV2$ is capturing the second-order statistics, \ $c_k, k \in [1, K]$ are the learnable clusters and  \ $\sigma_k, k \in [1, K]$ are the clusters' diagonal covariances.
To define $\sigma_k, k \in [1, K]$ as positive, we first randomly initialize their value with a Gaussian noise
with unit mean and small variance and then take the square of the values during training so that they stays
positive. In the same manner as NetVLAD, $c_{k}$ and $\sigma_{k}$ are learnt independently from the parameters of the soft-assignment $a_{k}$.
This formulation differs from~\cite{simonyan2013deepfisher,sydorov14deepfisher} as we are not exactly reproducing the original Fisher Vectors. Indeed the parameters $a_{k}(x_i),\ c_{k}$ and $\sigma_{k}$ are decoupled from each other. As opposed to~\cite{simonyan2013deepfisher,sydorov14deepfisher}, these parameters are not related to a Gaussian Mixture Model but instead are trained in a discriminative manner. 

Finally, we have also investigated a simplification of the original NetVLAD architecture that averages the actual descriptors instead of residuals, as first proposed by~\cite{douze13stable}. We call this variant NetRVLAD (for Residual-less VLAD).
This simplification requires less parameters and computing operations (about half in both cases). The NetRVLAD descriptor can be written as
\begin{align}
RVLAD(j,k) = \sum_{i=1}^{N} a_{k}(x_i)x_i(j).
\end{align}
More information about our Tensorflow~\cite{tensorflow} implementation of these different aggregation models can be found at: \url{https://github.com/antoine77340/LOUPE}


\section{Experiments} \label{exp}
This section evaluates alternative architectures for video aggregation and presents results on the Youtube-8M~\cite{youtube8m} dataset.

\subsection{Youtube-8M Dataset}

\begin{figure*}
  \begin{center}
     \includegraphics[width=\textwidth]{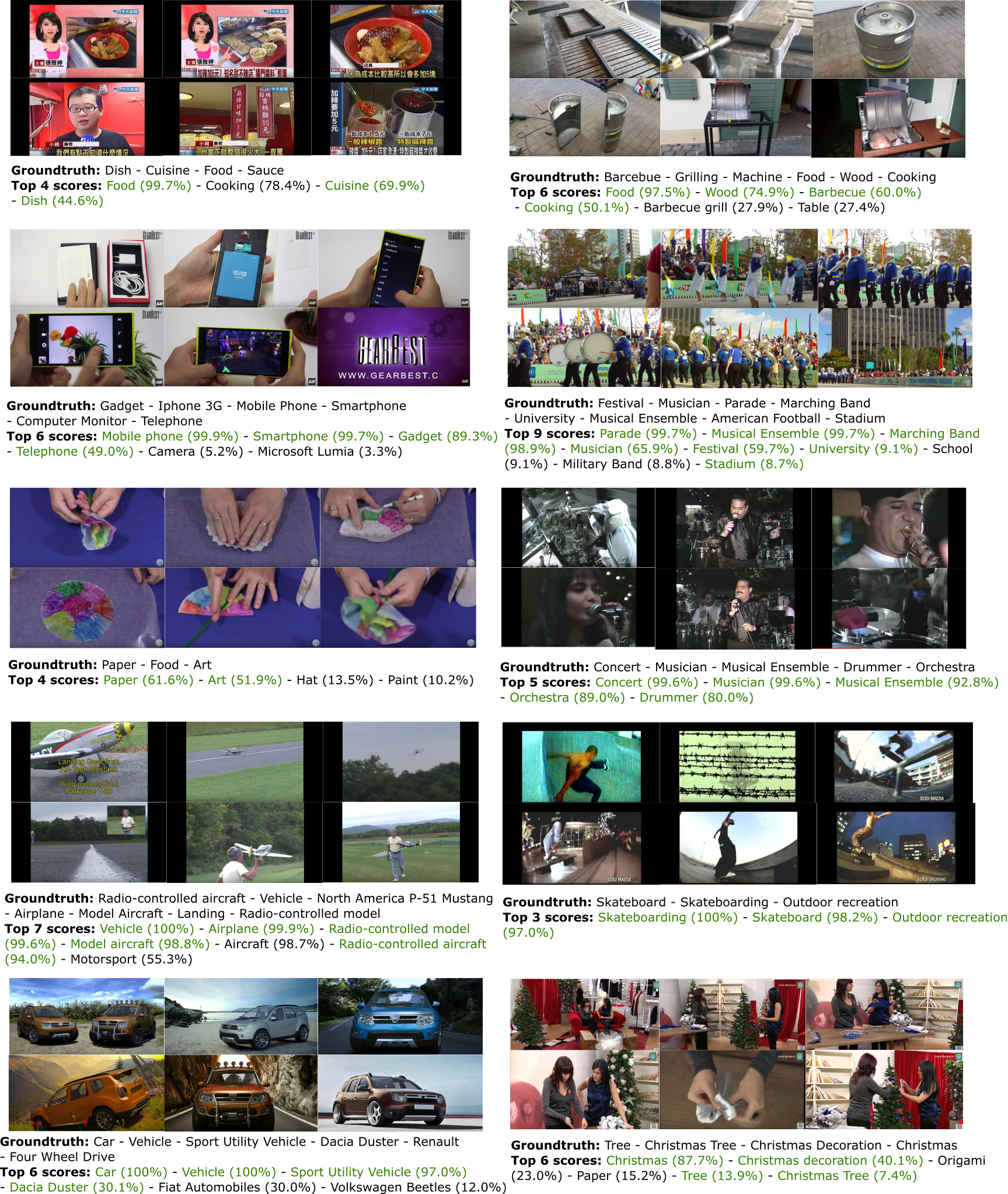}
\end{center}
  \caption{Qualitative results from our best single model (Gated NetVLAD). We show both groundtruth labels
  (in green) from the Youtube 8M dataset and the top predictions of the Gated NetVLAD model.}
\label{fig:qualit results}
\end{figure*}

The Youtube-8M dataset~\cite{youtube8m} is composed of approximately 8 millions videos.
Because of the large scale of the dataset, visual and audio features are pre-extracted and provided with the dataset.
Each video is labeled with one or multiple tags referring to the main topic of the video. Figure~\ref{fig:qualit results} illustrates examples of videos with their annotations. The original dataset is divided into training, validation and test subsets with 70\%, 20\% and 10\% of videos, respectively.
In this work we keep around 20K videos for the validation, the remaining samples from the original training and validation subsets are used for training.
This choice was made to obtain a larger training set and to decrease the validation time.
We have noticed that the performance on our validation set was comparable ($0.2$\%-$0.3$\% higher) to the test performance evaluated on the Kaggle platform.
As we have no access to the test labels, most results in this section are reported for our validation set.
We report evaluation using the Global Average Precision (GAP) metric at top 20 as used in the Youtube-8M Kaggle competition (more details about the metric can be found at: \url{https://www.kaggle.com/c/youtube8m#evaluation}).

\subsection{Implementation details}
In the Youtube 8M competition dataset~\cite{youtube8m} video and audio features are provided for every second of the input video. 
The visual features consist of ReLU activations of the last fully-connected layer from
a publicly available\footnote{\url{https://www.tensorflow.org/tutorials/image_recognition}} Inception network trained on Imagenet.
The audio features are extracted from a CNN architecture trained for audio classification~\cite{hershey17cnn}.
PCA and whitening are then applied to reduce the dimension to 1024 for the visual features and 128 for the audio features.
More details on feature extraction are available in~\cite{youtube8m}.

All of our models are trained using the Adam algorithm~\cite{kingma15adam} and mini-batches with data from around 100 videos.
The learning rate is initially set to $0.0002$ and is then decreased exponentially with the factor of $0.8$ every 4M samples.
We use gradient clipping and batch normalization~\cite{ioffe2015batchnorm} before each non-linear layer.

For the clustering-based pooling models, i.e.~BoW, NetVLAD, NetRVLAD and NetFV, we randomly sample $N$ features with replacement from each video.
$N$ is fixed for all videos at training and testing.
As opposed to the original version of NetVLAD~\cite{arandjelovic16netvlad}, we did not pre-train the codebook with a k-means initialization as we did not notice any improvement by doing so. 
For training of recurrent models, i.e.~LSTM and GRU, we process features in the temporal order.
We have also experimented with the random sampling of frames for LSTM and GRU which performs surprisingly similarly.

All our models are trained with the cross entropy loss. 
Our implementation uses the TensorFlow framework~\cite{tensorflow}.
Each training is performed on a single NVIDIA TITAN X (12Gb) GPU.

\begin{table}[t]
  \setlength{\tabcolsep}{3pt}
    \centering
    \scalebox{1}{
    \begin{tabular}{@{}lc@{}}
      \toprule
      Method                   & GAP\\
      \midrule
      Average pooling + Logistic Regression      & $71.4\%$ \\
      Average pooling + MoE + CG                 & $74.1\%$ \\
      \midrule
      LSTM (2 Layers)                & $81.7\%$ \\
      GRU (2 Layers)                 & $82.0\%$ \\
      \midrule
      BoW (4096 Clusters) & $81.6\%$ \\
      NetFV (128 Clusters)      & $82.2\%$ \\
      NetRVLAD (256 Clusters)          & $82.3\%$ \\
      NetVLAD (256 Clusters)          & $82.4\%$ \\
      \midrule
      Gated BoW (4096 Clusters) & $82.0\%$ \\
      Gated NetFV (128 Clusters)      & $83.0\%$ \\
      Gated NetRVLAD (256 Clusters)   & $83.1\%$ \\
      Gated NetVLAD (256 Clusters)    & $\textbf{83.2\%}$ \\
      \bottomrule
    \end{tabular}
    }
    \caption{
      Performance comparison for individual aggregation schemes. Clustering-based methods are compared with and without Context Gating.
    }
      \label{table:model-comparison}
\end{table}


\subsection{Model evaluation}
We evaluate the performance of individual models in Table~\ref{table:model-comparison}.
To enable a fair comparison, all pooled representations have the same size of 1024 dimensions.
The ``Gated'' versions for the clustering-based pooling methods include CG layers as described in Section~\ref{sec:cg}.
Using CG layers together with GRU and LSTM has decreased the performance in our experiments.

From Table~\ref{table:model-comparison} we can observe a significant increase of performance provided by all learnt aggregation schemes compared to the Average pooling baselines.
Interestingly, the NetVLAD and NetFV representations based on the temporally-shuffled feature pooling outperforms the temporal models (GRU and LSTM).
Finally, we can note a consistent increase in performance provided by the Context Gating for all clustering-based pooling methods. 

\subsection{Context Gating ablation study}
Table \ref{table:cg-comparison} reports an ablation study evaluating the effect of Context Gating on the NetVLAD aggregation with 128 clusters.
The addition of CG layers in the feature pooling and classification modules gives a significant increase in GAP.
We have observed a similar behavior for NetVLAD with 256 clusters.
We also experimented with replacing the Context Gating by the GLU~\cite{dauphin16GLU} after pooling. To make the comparison fair, we added
a Context Gating layer just after the MoE. Despite being less complex than GLU, we observe that CG also performs better.
We note that the improvement of $0.8$\% provided by CG is similar to the improvement of the best non-gated 
model (NetVLAD) over LSTM in Table~\ref{table:model-comparison}.


\subsection{Video-Audio fusion}

In addition to the late fusion of audio and video streams (Late Concat) described in Section~\ref{sec:arch}, we have also experimented with
a simple concatenation of original audio and video features into a single vector, followed by the pooling and classification modules in a ``single stream manner'' (Early Concat).
Results in Table~\ref{table:fusion-comparison} illustrate the effect of the two fusion schemes for different pooling methods.
The two-stream audio-visual architecture with the late fusion improves performance for the clustering-based pooling methods (NetVLAD and NetFV).
On the other hand, the early fusion scheme seems to work better for GRU and LSTM aggregations.
We have also experimented with replacing the concatenation fusion of audio-video features by their outer product. 
We found this did not work well compared to the concatenation mainly due to the high dimensionality of the resulting output. To alleviate this issue, we tried to 
reduce the output dimension using the multi-modal compact bilinear pooling approach~\cite{gao16compactbilinearpooling} but found the resulting models underfitting the data.

%


\begin{table}[t]
  \setlength{\tabcolsep}{3pt}
    \centering
    \scalebox{1}{
\begin{tabular}{@{}ccc@{}}
      \toprule
      After pooling & After MoE   & GAP\\
      \midrule
       - & -                             & $82.2\%$ \\
       Gated Linear Unit & -                            & $82.4\%$ \\
       Context Gating & -                             & $82.7\%$ \\
       Gated Linear Unit & Context Gating                           & $82.7\%$ \\
       Context Gating & Context Gating                       & $\textbf{83.0\%}$ \\
      \bottomrule
    \end{tabular}
     }
 \caption{Context Gating ablation study. There is no GLU layer after MoE as GLU does not output probabilities.}
      \label{table:cg-comparison}
\end{table}

\begin{table}[t]
  \setlength{\tabcolsep}{3pt}
    \centering
    \scalebox{1}{
    \begin{tabular}{@{}lcc@{}}
      \toprule
      Method                   & Early Concat & Late Concat\\
      \midrule
      NetVLAD                   & $81.9\%$ & $\textbf{82.4\%}$  \\
      NetFV                     & $81.2\%$ & $\textbf{82.2\%}$ \\
      GRU                       & $\textbf{82.2\%}$ & $82.1\%$ \\
      LSTM                      & $\textbf{81.7\%}$ & $81.1\%$ \\
      \bottomrule
          \end{tabular}
    }
   \caption{Evaluation of audio-video fusion methods (Early and Late Concat).}
      \label{table:fusion-comparison}
\end{table}

\subsection{Generalization}

\begin{figure}[t]
  \begin{center}
     \includegraphics[width=0.5\textwidth]{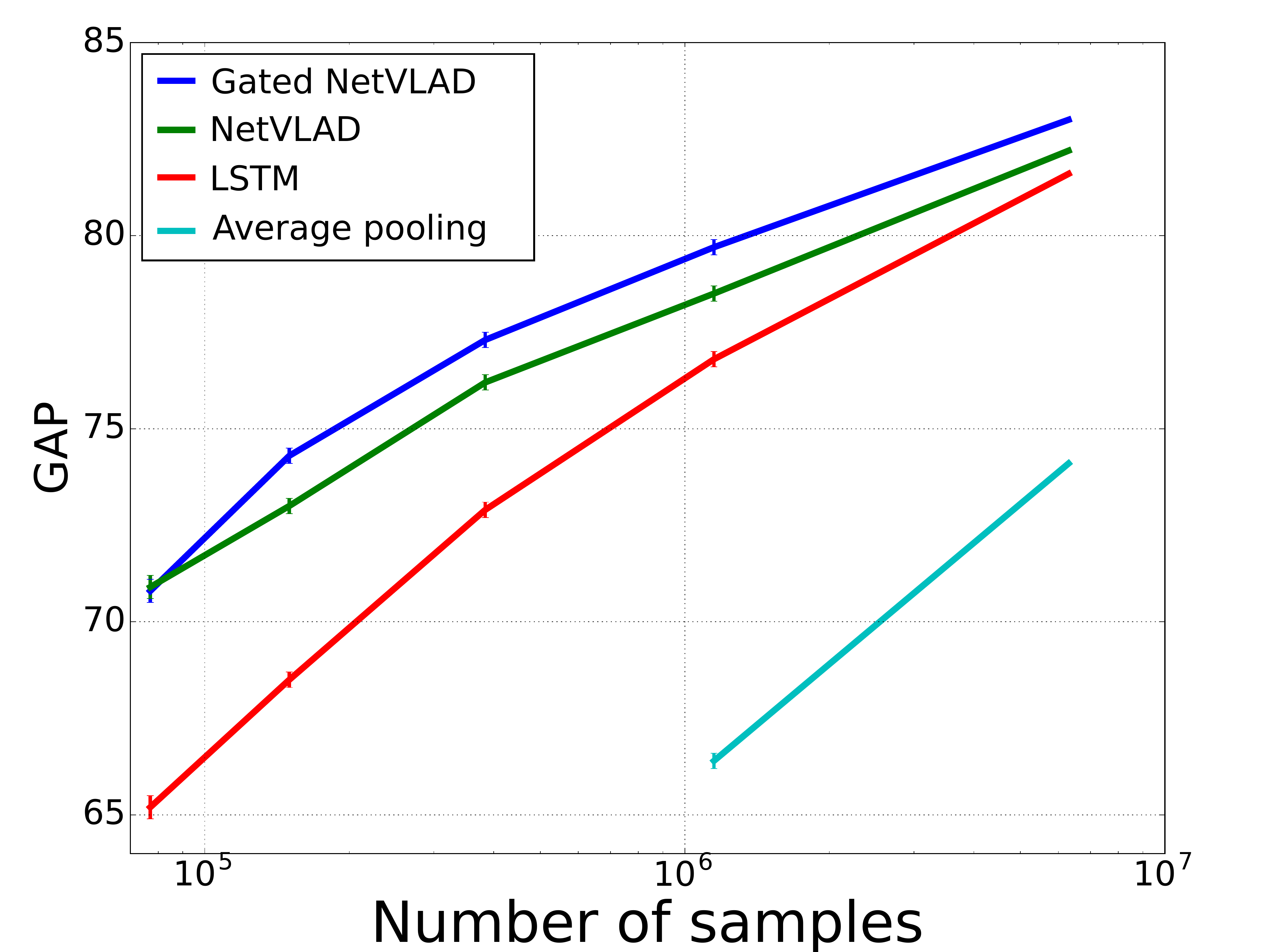}
\end{center}
  \caption{The GAP performance of the different main models when varying the
  dataset size.}
\label{fig:generalization}
\end{figure}

One valuable feature of the Youtube-8M dataset is the large scale of annotated data (almost 10 millions videos).
More common annotated video datasets usually have sizes several orders of magnitude lower, ranging from 10k to 100k samples.
With the large-scale dataset at hand we evaluate the influence of the amount of training data on the performance of different models.
To this end, we experimented with training different models: Gated NetVLAD, NetVLAD, LSTM and average pooling based model
on multiple randomly sampled subsets of the Youtube 8M dataset.
We have experimented with subsets of 70K, 150K, 380K and 1150K samples.
For each subset size,
we have trained models using three non-overlapping training subsets and measured the variance in performance.
Figure~\ref{fig:generalization} illustrates the GAP performance of each model when varying the training size. The error bars represent the variance observed when training the models on the three different training subsets.
We have observed low and consistent GAP variance for different models and training sizes.
Despite the LSTM model has less parameters (around 40M) compared to NetVLAD (around 160M) and Gated NetVLAD (around 180M), NetVLAD and Gated NetVLAD models demonstrate better generalization than LSTM when trained from a lower number of samples.
The Context Gating module still helps generalizing better the basic NetVLAD based architecture when having sufficient number of samples (at least 100k samples). 
We did not show results with smaller dataset sizes as the results for all models were drastically dropping down. This is mainly due to the fact that the task is a multi-label prediction problem with a large pool of roughly 5000 labels. As these labels have a long-tail distribution, decreasing the dataset size to less than 30k samples would leave many labels with no single training example. Thus, it would not be clear if the drop of performance is due to the aggregation technique or a lack of training samples for rare classes.

\subsection{Ensembling} \label{kaggle}

We explore the complementarity of different models and consider their combination through ensembling.
Our ensemble consists of several independently trained models.
The ensembling averages label prediction scores of selected models.
We have observed the increased effect of ensembling when combining diverse models. 
To choose models, we follow a simple greedy approach: we start with the best performing model and choose the next model by maximizing the GAP of the ensemble on the validation set.
Our final ensemble used in the Youtube 8M challenge contains 25 models.
A seven models ensemble is enough to reach the first place with a GAP on the private test set of 84.688.
These seven models correspond to: Gated NetVLAD (256 clusters), Gated NetFV (128 clusters), Gated BoW (4096 Clusters), BoW (8000 Clusters), Gated NetRVLAD (256 Clusters),
GRU (2 layers, hidden size: 1200) and LSTM (2 layers, hidden size: 1024).
Our code to reproduce this ensemble is available at:~\url{https://github.com/antoine77340/Youtube-8M-WILLOW}.
To obtain more diverse models for the final 25 ensemble, we also added all the non-Gated models, varied the number of clusters or varied the size of the pooled representation.

Table~\ref{table:ensemble-comparison} shows the ensemble size of the other top ranked approaches, out of 655 teams, from the Youtube-8M kaggle challenge. Besides showing the best performance at the competition, we also
designed a smaller set of models that ensemble more efficiently than others. Indeed, we need much less models in our ensemble than the other top performing approaches. Full ranking can be found at: \url{https://www.kaggle.com/c/youtube8m/leaderboard}.


\begin{table}[t]
  \setlength{\tabcolsep}{3pt}
    \centering
    \scalebox{1}{
    \begin{tabular}{@{}lcc@{}}
      \toprule
      Approach & Ensemble size & GAP\\
      \midrule
      Ours (Full) & 25 & 85.0\\
      Ours (Light) & 7 & 84.7\\
      \midrule
      Wang \textit{et al.}~\cite{youtube8m2nd} & 75 & 84.6 \\%
      Li \textit{et al.}~\cite{youtube8m3rd} & 57 & 84.5\\
      Chen \textit{et al.}~\cite{youtube8m4th} & 134 & 84.2 \\
      Skalic \textit{et al.}~\cite{youtube8m5th} & 75 & 84.2 \\
      \bottomrule
    \end{tabular}
    }
    \caption{
      Ensemble model sizes of the top ranked teams (out of 655) from the Youtube 8M kaggle competition.
   }
      \label{table:ensemble-comparison}
\end{table}

\section{Conclusions}
We have addressed the problem of large-scale video tagging and explored trainable variants of classical pooling methods (BoW, VLAD, FV) for
the temporal aggregation of audio and visual features.
In this context we have observed NetVLAD, NetFV and BoW to outperform more common temporal models such as LSTM and GRU.
We have also introduced the Context Gating mechanism and have shown its benefit for the trainable versions of BoW, VLAD and FV.
The ensemble of our individual models has been shown to improve the performance further, enabling our method to win the
Youtube 8M Large-Scale Video Understanding challenge.
Our TensorFlow toolbox LOUPE is available for download from~\cite{miech17loupe} and includes implementations of the Context Gating as well as learnable pooling modules used in this work.

\ifCLASSOPTIONcompsoc
  \section*{Acknowledgments}
\else
  \section*{Acknowledgment}
\fi

The authors would like to thank Jean-Baptiste Alayrac and Relja Arandjelović for valuable discussions
as well as the Google team for providing the Youtube-8M Tensorflow Starter Code.
This work has also been partly supported by ERC grants ACTIVIA (no.\
307574) and LEAP (no.\ 336845), CIFAR Learning in Machines $\&$ Brains
program, ESIF, OP Research, development and education Project IMPACT
No.\ CZ$.02.1.01/0.0/0.0/15\_003/0000468$ and a Google Research Award.

\ifCLASSOPTIONcaptionsoff
  \newpage
\fi



%
\bibliographystyle{IEEEtran}
\bibliography{IEEEabrv,master-biblio}

\end{document}